\documentclass[11pt]{article}
\usepackage[final]{acl}
\usepackage{tikz}
\usetikzlibrary{positioning, arrows.meta}
\usepackage{amsmath,amssymb,amsfonts}
\usepackage{graphicx}
\usepackage{booktabs}
\usepackage[T1]{fontenc}
\usepackage{times}
\usepackage{latexsym}
\usepackage[utf8]{inputenc}
\usepackage{microtype}
\usepackage{listings}
\usepackage{placeins}

\begin{document}
\makeatletter
\newcommand{\subfootnotesize}{\@setfontsize\subfootnotesize{7.5}{9}}
\makeatother


\title{Do LLMs Understand Context? A Knowledge Graph-Based Evaluation Framework}

\author{
  Subavarshana Arumugam$^{1*}$ \quad
  Mamta Nallaretnam$^{1*}$ \quad
  Kithuni Wickramasinghe$^{1*}$ \quad
  Chamath Gunapala$^{1*}$ \\
  Pragatheeswaran Vipulanandan$^{2}$ \quad
  Kamal Premaratne$^{2}$ \quad 
  Uthayasanker Thayasivam$^{2}$ \\[4pt]
  $^{1}$Department of Computer Science \& Engineering, University of Moratuwa, Sri Lanka \\
  $^{2}$Department of Electrical and Computer Engineering, University of Miami, USA \\
    \texttt{\{ subavarshanaa.21, nallaretnam.21, kithuni.21, chamathg.21\}@cse.mrt.ac.lk} \\
  \thanks{$^{*}$Equal contribution.}
  }

\maketitle

\begin{abstract}
While large language models (LLMs) have achieved remarkable linguistic capabilities, a profound question lingers at their core: do these models truly comprehend context or simply excel at pattern matching on an unprecedented scale? Contextual understanding in LLMs refers to the ability to correctly extract relevant information from a given context, integrate it into a coherent internal representation, and reason over it to produce factually consistent and contextually grounded responses. However, traditional methods such as BiLingual Evaluation Understudy (BLEU) and  perplexity simply measure surface-level performance. This reveals a critical gap in question answering (QA), where responses must be contextually grounded rather than simply being memorized associations. To fill this void, we propose a novel knowledge graph (KG) based evaluation framework for LLM contextual understanding in QA. Central to this is Semantic Structural Similarity for KGs (S3KG), a hybrid similarity measure combining structural and semantic signals into a single score. In addition, a diagnostic analysis framework is developed to identify and categorize reasoning errors at the triplet level, enabling fine-grained analysis of model failures. Together, across nine benchmarks, S3KG achieves F1 gains of up to $+7.6$ points over the strongest baseline and AUROC up to $0.973$.
\end{abstract}


\section{Introduction}


LLMs, such as Bidirectional Encoder Representations from Transformers (BERT) \citep{reimers2019sbert}, the Generative Pre-trained Transformer (GPT) models  \citep{radford2018improving}, and their advanced variants, have fundamentally transformed natural language processing (NLP). These models exhibit extraordinary proficiency in generating coherent, human-like text, answering complex questions, and executing a broad spectrum of tasks \citep{brown2020language}. However, a profound question lingers at the core of these models: do they genuinely understand the content they process or do they merely produce plausible outputs without true contextual comprehension \citep{zhu2024can}. Hallucination detection literature has largely bypassed this dimension, focusing instead on output-level signals such as semantic entropy and token sequence probabilities \citep{vipulanandan2026semantic}.

Traditional evaluation metrics---such as perplexity, BLEU \citep{bleu}, and other token-level matching methods on standardized benchmarks---primarily measure syntactic- or surface-level performance and fail to capture the depth of semantic comprehension or contextual understanding \citep{reiter2018bleu}. This limitation raises concerns about the reliability of LLMs in high-stakes scenarios such as QA systems in medicine and healthcare, defense, and legal analysis.


\section{Related Work}


Evaluating how LLMs understand and utilize contextual information remains a key challenge in QA systems, as fluent and plausible responses do not necessarily reflect faithful use of the provided context \citep{correctness2024faithfulness}. \citet{zhu2024can} evaluate LLMs across four core tasks---coreference resolution, discourse relation classification, dialogue state tracking, and query rewriting---showing that while they capture general contextual patterns, they fail to reveal which parts of the context are misunderstood, and their analysis does not extend to the QA domain. Complementing this, \citet{mesaqa} probe LLM reasoning by manipulating in-context examples, including logical modifications such as swapping ``AND''/``OR'', finding that models do not consistently obey formal reasoning rules nor exhibit clearly identifiable error patterns. Together, these works establish that current LLMs demonstrate strong surface-level comprehension yet still struggle with fine-grained contextual interpretation and logical reasoning. Evaluating these limitations in long-form LLM answers is particularly challenging, making the use of KGs which involve accurate extraction of relational triplets from text a promising evaluation approach.

KG construction has progressed from fixed-schema supervised pipelines to joint extraction architectures enabled by pretrained language models \citep{shang2022}. Embedding models such as TransE~\citep{bordes2013translating}, which represents relations as vector translations, and RotatE~\citep{sun2019rotate}, which models relations as complex-space rotations to capture symmetry and composition, makes them well-suited for link prediction on static KGs but ill-suited for cross-graph similarity. The Weisfeiler-Lehman (WL) kernel~\citep{shervashidze2011weisfeiler} compares graphs by iteratively aggregating neighbourhood labels into histograms, while the Wasserstein WL (WWL) kernel~\citep{togninalli2019wasserstein} replaces histogram comparison with Wasserstein distance for better handling of continuous attributes. Both treat node labels as opaque symbols, so semantically equivalent but syntactically distinct labels receive zero credit---a key limitation. \citet{kea} partially address this through SBERT-based  \citep{reimers2019sbert} semantic clustering with few-shot instruction tuning, yet the alignment remains lossy and falls short of a principled similarity measure for heterogeneous, independently constructed KGs. 


\section{Our Contributions}

Our work makes three main contributions. 
\begin{itemize}\itemsep0em
    \item \textbf{Semantic Structural Similarity for KGs (S3KG)} is a hybrid semantic-structural similarity metric that converts LLM responses and reference answers into KG triplets and produces a single interpretable evaluation score.
    \item \textbf{Contextual Understanding Score (CUS)} is a model-level aggregate of two complementary dimensions, factual accuracy (GoldSim) and contextual faithfulness (CtxSim), enabling cross-model comparison across benchmarks.
     \item \textbf{Triplet Analyzing Unit (TAU)} is a diagnostic component that identifies and classifies reasoning failures at the triplet level for fine-grained behavioral analysis of model outputs.
\end{itemize}



\section{Methodology}

\begin{figure}[h]
    \centering
    \includegraphics[width=\linewidth]{%
        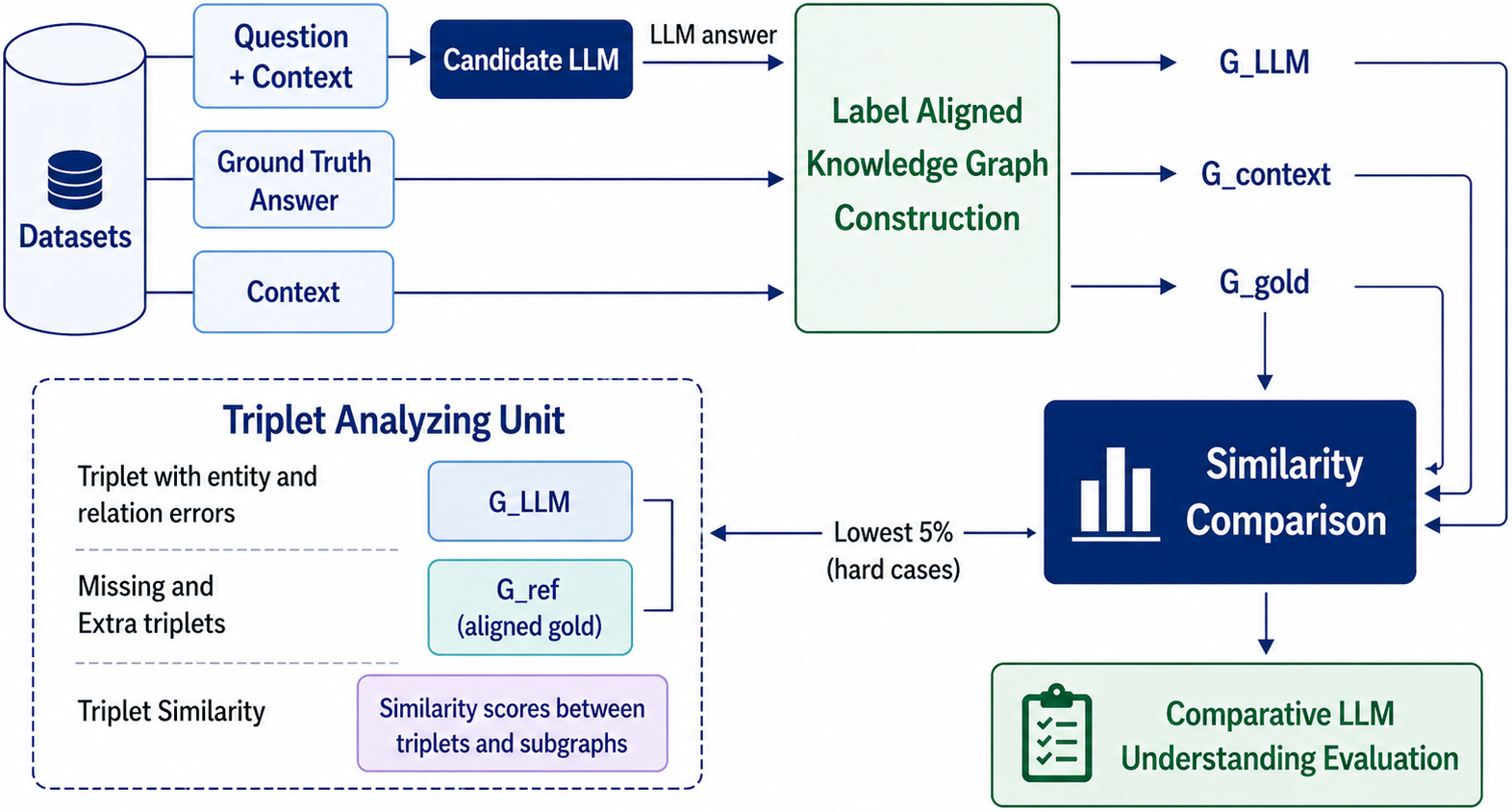}
    \caption{Methodology pipeline for LLM comparison and evaluation.}
    \label{fig:method}
\end{figure}

Our framework evaluates LLM contextual understanding using a KG-based pipeline (see Figure~\ref{fig:method}). Given a question and its supporting context, the LLM generates a response, from which KGs are constructed alongside those derived from the gold (or reference or ground truth) answer and context. These KGs are then compared to measure similarity. Low scoring pairs are further analyzed using a triplet analyzing unit to identify reasoning errors.

\subsection{LLM Answer Collection}
\label{sec:llm_collection}


Each model is prompted with a question paired with its associated supporting context, and the generated response is recorded alongside the gold answer to form the inputs for downstream evaluation. To establish a reproducible baseline, responses are first generated deterministically at temperature zero, yielding outputs that closely adhere to the provided context; subsequent runs are at progressively higher temperature settings allowing for us to examine how increasing generation diversity affects the model's ability to retain and utilize contextual information. These collected responses---together with the gold answers and supporting contexts---feed directly into the KG construction stage.


\subsection{KG Construction}
\label{sec:kg_construction}


For each QA instance, knowledge graphs are constructed from three sources: the gold answer, the model-generated response, and the supporting context. To ensure that the resulting KGs are comparable, we adopt the single few-shot prompting strategy with instruction tuning used in \citet{grapheval} and \citet{kea}, applying a shared extraction prompt uniformly across all three sources. This ensures that entities and relations are extracted under the same schema, encouraging a consistent entity and relation label space across all 3 KGs. Full details of the extraction prompt appear in Appendix~\ref{sec:llm-prompt}.

Following extraction, an additional NLP normalization step is applied uniformly to every entity and relation label across all 3 KGs. This includes lowercasing, lemmatization, and whitespace normalization, ensuring that any residual syntactic variation introduced during extraction is not retained in the final KG representations. Together, the consistency-aware prompting and post-extraction normalization ensure that the 3 KGs are structurally compatible and ready for meaningful comparison using S3KG.


\subsection{S3KG: Semantic Structural Similarity for KGs}
\label{sec:S3KG}


We denote a KG as $\mathcal{G} = \mathcal{G}(\mathcal{T})$, where $\mathcal{T} = \{(h_i, r_i, t_i),\; i \in \mathcal{I}\}$ is a collection of triplets enumerated via a finite index set $\mathcal{I}$. In the $i$-th triplet $(h_i, r_i, t_i)$, $h_i$, $r_i$, and $t_i$ are the head entity, relation, and tail entity, respectively. 

S3KG computes the similarity between two KGs $\mathcal{G}_1(\mathcal{T}_1)$ and $\mathcal{G}_2(\mathcal{T}_2)$. Here, for $k = 1, 2$, $\mathcal{T}_k = \{(h_{k,i}, r_{k,i}, t_{k,i}),\; i \in \mathcal{I}_k\}$. Rather than comparing KGs by a single criterion, S3KG operates at two levels: 
    (1)~a \textit{structural score} $\text{S}_{\text{WL}}$ computed node- and edge-wise using the WL kernel over soft-label aligned KGs, and 
    (2)~a \textit{semantic score} $\text{S}_{\text{SBERT}}$ computed triplet-wise from mean-pooled SBERT embeddings \citep{reimers2019sbert}.
These are blended via a mixing coefficient $\alpha$ to get the final combined similarity score (see \eqref{eq:S3KG}).


\textbf{Triplet-Level Matching.}
Triplets are serialised as natural language (NL) strings. SBERT embeddings are computed for all triplets in both sets $\mathcal{T}_1$ and $\mathcal{T}_2$ using \texttt{paraphrase-MPNet-base-v2}. For each triplet in $\mathcal{T}_1$, the most semantically similar triplet in $\mathcal{T}_2$ is selected by cosine similarity, producing a filtered set $\widehat{\mathcal{T}}_2 \subseteq \mathcal{T}_2$ that anchors the comparison to semantically relevant content. This unidirectional matching strategy is adopted deliberately. By treating $\mathcal{T}_1$ as the reference set, the method primarily evaluates how well the content of $\mathcal{T}_1$ is covered by $\mathcal{T}_2$, thereby emphasising recall while not penalising additional triplets present in $\mathcal{T}_2$. In contrast, a symmetric bidirectional formulation---computed by averaging matches from both $\widehat{\mathcal{T}}_2$ and $\widehat{\mathcal{T}}_1$, similar to the F1 formulation of BERTScore \citep{zhang2019bertscore}---would account for both recall and precision by also penalising unmatched surplus triplets. Investigating the empirical differences between the unidirectional and bidirectional variants is left for future work.


\textbf{Soft Label Alignment.}
The Standard WL kernel compares graphs by matching node labels exactly: two labels contribute to the similarity score only if they are syntactically identical strings. This means that semantically equivalent but syntactically different entity or relation labels (e.g., \textit{``found''} and \textit{``discovered''}) are treated as entirely distinct, and no credit is awarded for any semantic equivalence. S3KG resolves this through a soft label alignment step applied before kernel computation. Each node in $\mathcal{G}$ carries an entity label, and each edge carries a relation label. These are aligned independently to prevent cross-type collisions: for each entity label $\ell$ in $\mathcal{G}_1$, if its maximum cosine similarity, computed via SBERT embeddings $\phi(\cdot)$ to any entity label in $\mathcal{G}_2$ exceeds a threshold $\tau = 0.65$, it is mapped to the canonical identifier of the best-matching label in $\mathcal{G}_2$ (e.g., \texttt{node\_0}, \texttt{node\_1}); otherwise it is left unchanged. The same procedure is applied independently to relation labels (e.g., \texttt{rel\_0}, \texttt{rel\_1}). After alignment, syntactically different but semantically equivalent labels share the same canonical identifier, allowing the WL kernel to recognise them as matching.


\textbf{WL Kernel Structural Similarity.}
Following soft label alignment, the two KGs are compared using the WL graph kernel \citep{shervashidze2011weisfeiler}. The WL kernel operates iteratively: at iteration $0$, each node is characterised by its initial (soft aligned) entity label. At each subsequent iteration $k$, every node aggregates its current label with the multiset of its neighbours' labels and the connecting relation labels, producing a new refined label that encodes the node's $k$-hop neighbourhood structure. We use $K = 5$ iterations, so each node's final label summarises structural patterns up to 5 hops away. The kernel score is the normalised inner product between the resulting label-count histograms of the two graphs, yielding a structural similarity score $\text{S}_{\text{WL}} \in [0, 1]$ regardless of KG size.


\textbf{SBERT Mean-Pool Semantic Similarity.}
Each triplet $(h, r, t) \in \mathcal{T}$ is encoded by SBERT into an embedding $\phi(h,r,t)$; the graph-level representation is the mean $\bar{\mathbf{e}}_{\mathcal{T}} = \dfrac{1}{|\mathcal{T}|}\sum_{\mathcal{T}} \phi(h,r,t)$. Semantic similarity is the cosine between the mean-pooled representations of $\mathcal{T}_1$ and $\widehat{\mathcal{T}}_2$ (the semantically filtered reference triplets from Step 1), clipped to $[0,1]$:
\begin{equation}
  \text{S}_{\text{SBERT}}(\mathcal{T}_1,\, \widehat{\mathcal{T}}_2) 
    = \max
      \left(
        0, 
        \frac{%
        \bar{\mathbf{e}}_{\mathcal{T}_1} 
        \cdot
        \bar{\mathbf{e}}_{\widehat{\mathcal{T}}_2}}{%
        \|\bar{\mathbf{e}}_{\mathcal{T}_1}\|
        \cdot
        \|\bar{\mathbf{e}}_{\widehat{\mathcal{T}}_2}\|}
      \right).
\label{eq:sbert-score}
\end{equation}
This captures sentence-level meaning that discrete WL label refinement cannot.


\textbf{Combined Score.}
The structural and semantic scores are combined via mixing coefficient $\alpha$ as
\begin{equation}
  \text{S}_{\text{S3KG}} 
    = (1-\alpha)\, \text{S}_{\text{WL}} + \alpha\, \text{S}_{\text{SBERT}}.\;
      \alpha \in [0, 1].
\label{eq:S3KG}
\end{equation}
We use $\alpha = 0.5$ so that the score equally weights structural fidelity from WL neighbourhood aggregation over aligned labels and semantic coherence from SBERT mean-pooling, thus encoding both local relational patterns and global meaning within a single score.


\subsection{Contextual Understanding Score (CUS)}
\label{sec:cus}


With the 3 KGs---$\mathit{KG}_{\text{LLM}}$  associated with the model-generated response, $\mathit{KG}_{\text{gold}}$ associated with the gold answer, and $\mathit{KG}_{\text{ctx}}$ associated with the supporting context---in hand, for each QA instance $q$, we apply S3KG to get 
    (1)~$\mathrm{GoldSim}(q) = \text{S}_{\text{S3KG}}(\mathit{KG}_{\text{LLM}}^{(q)}, \mathit{KG}_{\text{gold}}^{(q)})$ which measures factual accuracy by comparing the LLM response against the gold answer; and 
    (2)~$\mathrm{CtxSim}(q) = \text{S}_{\text{S3KG}}(\mathit{KG}_{\text{LLM}}^{(q)}, \mathit{KG}_{\text{ctx}}^{(q)})$ which measures contextual faithfulness by comparing the LLM response against the supporting context. Since neither dimension alone reflects true understanding, we employ the harmonic mean to generate a \emph{Contextual Understanding Score (CUS)} as
\begin{equation}
  \mathrm{CUS}(q) 
    = \frac{%
      2 
      \cdot 
      \mathrm{GoldSim}(q) 
      \cdot 
      \mathrm{CtxSim}(q)}{%
      \mathrm{GoldSim}(q) + \mathrm{CtxSim}(q)}.
\label{eq:cus_sample}
\end{equation}
This harmonic mean penalises imbalanced profiles, ranking a model having one strong and one weak score below one having a pair of moderate scores. The dataset-level CUS is the mean of $\mathrm{CUS}(q)$ over all $N$ samples.


\subsection{Triplet Analysis Unit (TAU)}
\label{sec:triplet_unit}


For the 5\% of lowest-scoring QA pairs, we apply a triplet analysis unit (TAU) to identify where and how the LLM generated KG diverges from the gold KG. Each triplet is converted into an NL sentence and encoded using a sentence transformer model. Cosine similarity is computed between gold  and LLM triplet embeddings; aligned triplets are identified by thresholding the cosine similarity between triplet sentence embeddings.

After removing aligned triplets, residual pairs are categorized into interpretable error classes based on component-wise cosine similarities for head, relation, and tail: 
    (1)~\textit{Relation mismatch}: entities match, but the relation differs. 
    (2)~\textit{Entity mismatch}: the relation aligns, but the entity pair is inconsistent. 
    (3)~\textit{Extra triplets}: hallucinated or additional triplets generated by the LLM. 
    (4)~\textit{Missing triplets}: relevant triplets that were not extracted by the LLM. Full TAU evaluation details appear in  Appendix~\ref{sec:diagnosing}.


\section{Experiments}
\label{sec:experiments}


\subsection{Datasets}
\label{sec:datasets}


Two context-rich QA datasets containing long-form answers are used for evaluation purposes.     
    (1)~\textbf{PubMedQA} \citep{jin2019pubmedqa} contains 273,518 biomedical QA pairs drawn from research articles, with answers typically exceeding 100 words, providing a testbed for domain-specific detailed response evaluation. 
    (2)~\textbf{MesaQA} \citep{wang-etal-2025-mesaqa} comprises approximately 6,100 QA pairs from consumer healthcare documents, featuring abstractive answers averaging 70 words that require multi-span evidence integration. 
Together, these two datasets provide evaluation coverage across academic biomedical reasoning and practical healthcare knowledge synthesis.


\subsection{LLM Answer Collection}
\label{sec:exp_llm_collection}


We evaluate 4 instruction-tuned open-source language models with 7-billion parameters: 
    \texttt{Llama-2-7b-chat-hf}, 
    \texttt{Gemma-7b-it}, 
    \texttt{Mistral-7B-Instruct-v0.2}, and 
    \texttt{Falcon-7B-Instruct}. These specific models are selected because they operate at a similar scale, thus allowing for a fair comparison without the influence of model size. Each model is prompted with a question and its associated supporting context, and the model-generated response is recorded alongside the gold answer for evaluation. To establish a baseline, responses are first generated deterministically at $0$ temperature, yielding outputs that closely adhere to the provided context. Additional responses are then generated at higher temperature settings (we use $0.3$, $0.7$, and $1.0$) to examine how increasing generation diversity influences the model's ability to retain and utilize contextual information. Full temperature analysis appears in Appendix~\ref{sec:appendix_temp}.


\subsection{Benchmarking Dataset Collection}


To assess generalisation across diverse text types, we use 10 datasets spanning three structural categories. Each dataset is cast as a binary classification task: given a text pair $(s_1, s_2)$, predict whether the pair is semantically equivalent (label $= 1$) or not (label $= 0$). All datasets are balanced at $N$ positive and $N$ negative pairs (we use $N = 400$). Performance is measured by maximum F1 score (obtained via threshold sweep) and AUROC.


\textbf{Short-Text with Human-Annotated.} 
We use 3 datasets containing sentence pairs averaging 10--22 words with crowd-sourced or expert equivalence labels:   
    \textbf{MRPC} \citep{dolan2005mrpc}, consisting of news sentence pairs with paraphrase labels; \textbf{PAWS-Wiki} \citep{zhang2019paws}, adversarially constructed paraphrase pairs from Wikipedia where lexical overlap is deliberately an unreliable signal; and
    \textbf{STS12} \citep{agirre2012semeval}, sentence similarity pairs drawn from multiple NLP tasks.


\textbf{KG-Perturbed Paragraphs.} 
We use 6 datasets containing paragraphs averaging 69--126 words, constructed by perturbing entity relationships in KG-derived paragraph representations. The 5 evaluated datasets---\textbf{SK-Codex~400},
\textbf{SK-Combined}, \textbf{SK-FindKG}, \textbf{SK-GloBI}, and \textbf{SK-Oregano}---differ in their underlying KG source, covering general encyclopaedic (Codex \citep{safavi2021codex}), financial/economic (FindKG \citep{findkg}), biological interaction (GloBI \citep{globiref}), and food ontology (Oregano \cite{oregano}) domains.


\textbf{NLP-Perturbed Paragraphs.} 
\textbf{NLP-Perturbed Paragraphs.}
The \textbf{Wikipedia Entity-Swap} dataset (399 pairs) replaces named entities in Wikipedia passages using four NLP-based perturbations---node replacement, node deletion, edge deletion, and edge replacement via WordNet antonyms~\cite{miller1992wordnet}---with no KG involvement at any stage~\cite{sennrich-etal-2016-neural, wei-zou-2019-eda}. It serves as an anti-circularity probe: were our gains an artifact of circular evaluation, performance here should collapse, but it does not.


\subsection{Methods Evaluated}
\label{sec:methods}


\begin{table*}[t]
    \centering
    \caption{Similarity scores on two PAWS-Wiki pairs (threshold $= 0.5$;\checkmark\,=\,correct, $\times$\,=\,incorrect).The positive pair differs only in word order; the negative pair swaps the subject and object of the winning relation. All seven baselines assign near-identical high scores to both pairs, failing on the negative example. For the \textbf{positive pair}, S3KG achieves the optimal score of $1.00$, whereas surface-form methods such as BLEU ($0.58$) and ROUGE-L ($0.85$) underperform by penalising inconsequential word-order variation.For the \textbf{negative pair}, S3KG scores $0.48$---the only sub-threshold result---correctly predicting \textit{Not Similar}. The score is not zero because the sentences share substantial content; only the relational direction differs. The KG component isolates this reversal via the directed triplet, reducing the score from the near-$1.0$ surface baseline to just below the decision threshold, while $\alpha=0.5$ balances surface and structural similarity.}
    \label{tab:worked_example}
    \setlength{\tabcolsep}{5pt}
    \begin{tabular}{l p{6.1cm} p{6.1cm}}
        \toprule
        {}
            & \textbf{Positive Example} 
            & \textbf{Negative Example} \\
        \midrule
        \textbf{Text $s_1$} 
            & \textit{His father returned as a finished violinist of the Russian School to Bombay.} 
            & \textit{Renzo Furlan won 6--3, 6--4 against Thomas Johansson in the finals.} \\[4pt]
        \textbf{Text $s_2$} 
            & \textit{His father returned to Bombay as a finished violinist of the Russian school.} 
            & \textit{Thomas Johansson won 6--3, 6--4 against Renzo Furlan in the finals.} \\[4pt]
        \textbf{Triplet ($s_1$)} 
            & \small{(father, returned\_to, Bombay)} 
            & \small{(Renzo Furlan, won\_against, Thomas Johansson)} \\[2pt]
        \textbf{Triplet ($s_2$)} 
            & \small{(father, returned\_to, Bombay)} 
            & \small{(Thomas Johansson, won\_against, Renzo Furlan)} \\
        \midrule
        \textbf{True Label} 
            & \hfil\textbf{1} (Similar) 
            & \hfil\textbf{0} (Not Similar) \\
        \midrule
        \multicolumn{3}{l}{%
        \textit{Similarity scores (\checkmark \,=\,  correct prediction; $\times$ \,=\, incorrect prediction)}} \\[3pt]
        \textbf{S3KG} (Ours) 
            & \hfil\textbf{1.00} \checkmark  
            & \hfil\textbf{0.48} \checkmark \\
        ROUGE-1 
            & \hfil1.00 \checkmark           
            & \hfil1.00 $\times$ \\
        ROUGE-2             
            & \hfil0.83 \checkmark           
            & \hfil0.75 $\times$ \\
        ROUGE-L
            & \hfil0.85 \checkmark           
            & \hfil0.69 $\times$ \\
        BLEU
            & \hfil0.58 \checkmark           
            & \hfil0.70 $\times$ \\
        BERTScore
            & \hfil0.98 \checkmark           
            & \hfil0.98 $\times$ \\
        MiniLM
            & \hfil1.00 \checkmark           
            & \hfil0.94 $\times$ \\
        sentence-T5-base
            & \hfil1.00 \checkmark           
            & \hfil0.99 $\times$ \\
        \bottomrule
    \end{tabular}
\end{table*}

\textbf{S3KG} is evaluated with a mixing coefficient $\alpha$ (see  \eqref{eq:S3KG}) which controls the blend between KG structural signal and sentence-transformer signal: $\alpha = 0.0$ recovers a pure KG structural embedding ; $\alpha = 1.0$ recovers a pure dense sentence-transformer representation. Pure KG structural embeddings capture relational and ontological structure between concepts but lack linguistic flexibility and contextual expressiveness, whereas sentence transformers excel at contextual and semantic similarity yet remain blind to the underlying KG topology. S3KG bridges this gap by interpolating between both signals, enabling richer matching that is sensitive to both conceptual structure and NL meaning. For each dataset, we report the best-performing variant selected by maximum F1 across the sweep $\alpha \in \{0.0, 0.1, \ldots, 1.0\}$. Results for a more complete per-dataset $\alpha$ sweep appear in Appendix~\ref{app:alpha_sweep}. We compare against 7 standard baselines: ROUGE-1, ROUGE-2, ROUGE-L \citep{lin2004rouge}, BLEU \citep{bleu}, BERTScore \citep{zhang2019bertscore}, MiniLM \citep{wang2020minilm}, and sentence-T5-base \citep{ni2022t5}. Table~\ref{tab:worked_example} provides a concrete worked example demonstrating how S3KG detects relational reversals that all 7 baselines fail to distinguish.

All experiments\footnote{ are available at: \href{https://github.com/aaivu/knowledge-xtraction}{https://github.com/aaivu/knowledge-xtraction}} and were conducted on a workstation equipped with 64 \, GB RAM and an NVIDIA A6000 GPU.


\section{Results}
\label{sec:results}


We evaluate S3KG against seven baselines on 9 benchmarks spanning short-text paraphrase detection, KG-perturbed paragraphs, and an anti-circularity entity-swap control. Table~\ref{tab:heatmap} summarises F1 and AUROC across every dataset $\times$ method cell; the per-dataset best $\alpha$ together with the headline scores appear in Table~\ref{tab:summary}. Full per-dataset performance tables (short-text, KG-perturbed paragraph, Wikipedia entity-swap) appear in Appendix~\ref{app:per_dataset}; the complete $\alpha$ sweep appears in Appendix~\ref{app:alpha_sweep}.


\begin{table*}[!t]
    \centering
    \caption{F1 Score and ROC-AUC of S3KG and seven baselines across all nine benchmark datasets. 
    The best-performing method per dataset is \textbf{bolded}. 
    S3KG is shown using the best $\alpha$ variant per dataset, selected by maximum F1 over $\alpha$ swept over $\alpha \in \{0.0, 0.1, \ldots, 1.0\}$ (full sweep in Appendix~\ref{app:alpha_sweep}).
    S3KG attains the highest score on 6 of 9 datasets and the highest \emph{meaningful} score on the Wikipedia Entity-Swap anti-circularity control, with gains of up to $+7.6$ F1 over the strongest baseline on KG-rich paragraph datasets.}
    \label{tab:heatmap}
    \setlength{\tabcolsep}{3pt}
    \renewcommand{\arraystretch}{1.1}
    \small
    \smallskip
    \textbf{F1 Score}
    \smallskip
    \resizebox{\textwidth}{!}{%
    \begin{tabular}{l cccccccc}
        \toprule
        \textbf{Dataset} 
            & \textbf{S3KG (Ours)} & \textbf{ROUGE-1} & \textbf{ROUGE-2} & \textbf{ROUGE-L} & \textbf{BLEU} & \textbf{BERTScore} & \textbf{MiniLM} & \textbf{sent-T5-base} \\
        \midrule
        MRPC
            & 0.692 & 0.745 & 0.720 & 0.729 & 0.687 & 0.758 & 0.723 & \textbf{0.765} \\
        PAWS-Wiki
            & \textbf{0.766} & 0.678 & 0.715 & 0.735 & 0.716 & 0.691 & 0.687 & 0.674 \\
        Semantic-KG Combined  
            & \textbf{0.834} & 0.732 & 0.707 & 0.722 & 0.715 & 0.757 & 0.770 & 0.774 \\
        Wiki Swap
            & 0.872 & \textbf{1.000} & 0.860 & 0.729 & 0.868 & 0.747 & 0.821 & 0.767 \\
        Semantic-KG Codex 400 
            & \textbf{0.932} & 0.835 & 0.822 & 0.792 & 0.806 & 0.823 & 0.875 & 0.872 \\
        Semantic-KG FindKG    
            & 0.767 & 0.745 & 0.717 & 0.719 & 0.711 & 0.739 & 0.802 & \textbf{0.848} \\
        Semantic-KG GloBI     
            & \textbf{0.892} & 0.784 & 0.776 & 0.763 & 0.775 & 0.816 & 0.780 & 0.730 \\
        Semantic-KG Oregano   
            & \textbf{0.812} & 0.745 & 0.752 & 0.792 & 0.745 & 0.743 & 0.773 & 0.800 \\
        STS12
            & 0.786 & 0.725 & 0.681 & 0.703 & 0.671 & 0.682 & 0.833 & \textbf{0.856} \\
        \bottomrule
    \end{tabular}%
    }
    \textbf{AUROC}
    \smallskip
    \resizebox{\textwidth}{!}{%
    \begin{tabular}{l cccccccc}
        \toprule
        \textbf{Dataset} 
            & \textbf{S3KG (Ours)} & \textbf{ROUGE-1} & \textbf{ROUGE-2} & \textbf{ROUGE-L} & \textbf{BLEU} & \textbf{BERTScore} & \textbf{MiniLM} & \textbf{sent-T5-base} \\
        \midrule
        MRPC
            & 0.673 & 0.784 & 0.721 & 0.760 & 0.677 & \textbf{0.816} & 0.748 & \textbf{0.816} \\
        PAWS-Wiki
            & 0.795 & 0.490 & 0.721 & \textbf{0.807} & 0.747 & 0.702 & 0.638 & 0.668 \\
        Semantic-KG Combined  
            & \textbf{0.829} & 0.728 & 0.711 & 0.717 & 0.708 & 0.792 & 0.789 & 0.828 \\
        Wiki Swap
            & \textbf{0.890} & 1.000 & 0.772 & 0.311 & 0.745 & 0.645 & 0.811 & 0.790 \\
        Semantic-KG Codex 400 
            & \textbf{0.973} & 0.917 & 0.894 & 0.855 & 0.884 & 0.916 & 0.943 & 0.945 \\
        Semantic-KG FindKG    
            & 0.796 & 0.745 & 0.706 & 0.721          & 0.710 & 0.761 & 0.844 & \textbf{0.902} \\
        Semantic-KG GloBI
            & \textbf{0.935} & 0.833 & 0.833 & 0.800 & 0.819 & 0.871 & 0.817 & 0.761 \\
        Semantic-KG Oregano   
            & \textbf{0.892} & 0.782 & 0.791 & 0.835 & 0.794 & 0.798 & 0.814 & 0.871 \\
        STS12
            & 0.834 & 0.754 & 0.656 & 0.710 & 0.644 & 0.636 & 0.894 & \textbf{0.928} \\
        \bottomrule
    \end{tabular}%
    }
\end{table*}

\begin{table}[t]
    \centering
    \caption{Best S3KG variant per dataset, selected by maximum F1 via grid search over $\alpha \in \{0.0, 0.1, \ldots, 1.0\}$. 
    KG-rich paragraph datasets gain up to $+7.6$ F1 over the strongest baseline; 
    sparse or noisy KGs yield reduced margins.}
    \label{tab:summary}
    \begin{tabular}{l c c c}
    \toprule
    \textbf{Dataset} 
        & \textbf{Best $\alpha$} 
        & \textbf{F1} 
        & \textbf{AUROC} \\
    \midrule
    MRPC
        & 0.3 & 0.692 & 0.673 \\
    PAWS-Wiki
        & 0.5 & 0.766 & 0.795 \\
    STS12
        & 0.1 & 0.786 & 0.834 \\
    SK-Codex 400 
        & 0.5 & 0.932 & 0.973 \\
    SK-Combined  
        & 0.5 & 0.834 & 0.829 \\
    SK-FindKG    
        & 0.0 & 0.767 & 0.796 \\
    SK-GloBI     
        & 0.6 & 0.892 & 0.935 \\
    SK-Oregano   
        & 0.4 & 0.812 & 0.892 \\
    Wiki Swap    
        & 0.1 & 0.872 & 0.890 \\
    \bottomrule
\end{tabular}
\end{table}


\textbf{Text Richness Drives KG Performance.} 
S3KG performance scales with text length and relational density. On short texts (MRPC, STS12; 10--22 words average), the structural KG signal is sparse because only a few well-formed triplets can be extracted, and S3KG is competitive but trails sentence-T5-base by 6--7 F1 points. On paragraph-level datasets (69--126 words), S3KG reaches top-1 performance on 4 of 5 KG-perturbed benchmarks---SK-Codex~400 (F1 $= 0.932$, $+5.7$ over MiniLM), SK-Combined ($0.834$, $+6.4$ over sentence-T5-base), SK-GloBI ($0.892$, $+7.6$ over BERTScore), and SK-Oregano ($0.812$, $+1.2$ over sentence-T5-base)---confirming that sufficient relational content is required for the structural channel to pay off. On the adversarial PAWS-Wiki paraphrase benchmark, S3KG still leads (F1 $= 0.766$) while ROUGE-1 collapses to near-random (AUC $= 0.490$), reflecting the well-known failure of $n$-gram overlap on surface-form-preserving rephrasings.


\textbf{KG Extraction Quality is a Bottleneck.} 
On SK-FindKG, where financial and economic vocabulary degrades triplet extraction quality, sentence-T5-base leads (F1 $= 0.848$) and S3KG drops to $0.767$. The best $\alpha$ for this dataset is $0.0$ (pure KG), but the absolute score remains capped by noisy triplets, evidence that S3KG's gains depend on the underlying graph being faithfully recoverable from text.


\textbf{Anti-Circularity Validation.} 
On the Wikipedia Entity-Swap control, which is constructed independently of any KG used during S3KG development, S3KG attains the best meaningful score
(F1 $= 0.872$, AUC $= 0.890$), exceeding MiniLM (F1 $= 0.821$) and sentence-T5-base (F1 $= 0.762$). ROUGE-1 is excluded from the meaningful comparison because entity-swapped pairs share nearly all surrounding tokens, making unigram overlap trivially near-perfect, a dataset artifact rather than a real signal. The result rules out circularity as an explanation for S3KG's KG-perturbed gains.


\textbf{Optimal $\alpha$ is Dataset-Dependent.} 
Lower $\alpha$ values favour datasets where perturbations are primarily structural (SK-FindKG: $\alpha = 0.0$; STS12, Wiki Swap: $\alpha = 0.1$). Higher values are preferred when KG and dense signals are complementary (SK-Codex~400 and SK-Combined: $\alpha = 0.5$; SK-GloBI: $\alpha = 0.6$). The full sweep appears in Appendix~\ref{app:alpha_sweep}.


\textbf{Baseline Behaviour.} 
Token-overlap baselines (ROUGE, BLEU) are strong on KG-perturbed datasets where perturbations alter surface form, but unreliable on adversarial datasets (PAWS-Wiki: ROUGE-1 AUC $= 0.490$; Wiki Swap: ROUGE-L AUC $= 0.311$). BERTScore is more stable but consistently underperforms S3KG on KG-perturbed data. MiniLM and sentence-T5-base are the strongest baselines overall but require full fine-tuned transformer inference, whereas S3KG's KG component is comparatively lightweight at inference time.


\subsection{LLM Comparison on QA Datasets}
\label{sec:llm_comparison}


Having validated S3KG as a reliable KG similarity measure, we apply it as an evaluation instrument to address the following question: \textit{given a question and its supporting context, to what extent does an LLM capture the relational knowledge of the reference answer, and how faithfully does its response reflect the provided context?} Traditional metrics such as BLEU or exact match are insufficient for this purpose, as they assess syntactic-level token overlap rather than the relational knowledge structure of a response.

Using the CUS evaluation pipeline in Section~\ref{sec:cus}, Table~\ref{tab:llm} reports mean GoldSim, CtxSim, and CUS for four 7B-parameter models on PubMedQA and MesaQA ($N = 400$,  $\alpha = 0.5$).

\begin{table}[t]
    \centering
    \caption{LLM Evaluation Results (mean over $N{=}400$ samples,, $\alpha{=}0.5$). GoldSim measures factual alignment with the reference answer; CtxSim measures faithfulness to the supporting context; CUS (Equation~\ref{eq:cus_sample}) is their harmonic mean, penalising imbalanced profiles. Mistral-7B achieves the best CUS on both datasets, reflecting consistently balanced factual and contextual understanding, while Falcon-7B underperforms across all metrics and both domains.}
    \label{tab:llm}
    \resizebox{\columnwidth}{!}{%
    \begin{tabular}{l l ccc}
        \toprule
        \textbf{Dataset} 
            & \textbf{Model} 
            & \textbf{GoldSim} & \textbf{CtxSim} & \textbf{CUS} \\
        \midrule
        {}
            & Gemma-7B   
            & \textbf{0.6889} & 0.7031 & 0.6774 \\
        MesaQA   
            & Llama-2-7B 
            & 0.6570 & 0.7236 & 0.6752 \\
        {}
            & Mistral-7B 
            & 0.6491 & \textbf{0.7356} & \textbf{0.6780} \\
        {}
            & Falcon-7B  
            & 0.6036 & 0.6458 & 0.6063 \\
        \midrule
        {}
            & Gemma-7B   
            & \textbf{0.5235} & 0.6401 & 0.5587 \\
        PubMedQA 
            & Llama-2-7B 
            & 0.5220 & 0.6567 & 0.5651 \\
        {}
            & Mistral-7B 
            & 0.5138 & \textbf{0.7331} & \textbf{0.5923} \\
        {}
            & Falcon-7B  
            & 0.4541 & 0.5560 & 0.4800 \\
        \bottomrule
    \end{tabular}%
    }
\end{table}


\textbf{MesaQA Results.}
Gemma-7B, Llama-2-7B, and Mistral-7B achieve similar CUS scores ($0.675$--$0.678$), while Falcon-7B scores notably lower ($0.606$). Gemma-7B leads on GoldSim ($0.689$), while Mistral-7B leads on CtxSim ($0.736$) and achieves the best overall CUS ($0.678$) by balancing both dimensions. CtxSim consistently exceeds GoldSim across all models, indicating that models draw effectively from context but add content not present in the reference answer.


\textbf{PubMedQA Results.}
Performance drops substantially, with CUS ranging from $0.480$ (Falcon-7B) to $0.592$ (Mistral-7B), reflecting the difficulty of matching precise biomedical reference answers. Mistral-7B again leads in CUS, supported by the highest CtxSim ($0.733$). Gemma-7B scores highest on GoldSim ($0.524$) but lowest on CtxSim ($0.640$), indicating closer alignment with reference content but weaker use of biomedical context. Falcon-7B is weakest across all metrics on both datasets.


\textbf{Cross-Dataset Observations.}
CtxSim exceeds GoldSim in all 8 model--dataset combinations, consistent with instruction-tuned models elaborating on context rather than producing concise reference-style responses. The $\sim$10-point CUS gap between MesaQA and PubMedQA across all models points to domain complexity as the primary factor, with biomedical vocabulary and reasoning posing challenges irrespective of model architecture.


\textbf{Triplet Analysis of the Least Similar KGs.} 
Across the 5\% lowest-similarity cases in both datasets, the four models show clear differences, and we focus on this tail subset because these challenging instances make model failures more diagnostic and reveal systematic weaknesses that can be masked by strong average scores. Mistral is consistently the most reliable on the hard examples, preserving more reference facts and producing more aligned triplets than the other models. Llama generally falls in the middle, while Gemma shows the weakest performance, especially on the general health QA set where it often fails to recover many gold-aligned facts. A second consistent pattern is that, in these difficult cases, the generated KGs tend to align more closely with the contextual KG than with the reference KG. This suggests that many failures are not simply random errors, but cases where the model either drifts toward a different interpretation, omits key reference facts, or introduces unsupported additions. Finally, the PubMed setting is noticeably harder for all models: aligned-triplet recovery drops sharply, indicating that technical biomedical terminology, abbreviations, and entity variability dominate the failure modes in the hardest cases.


\section{Discussion}


The proposed KG-based framework provides a structured approach to evaluate LLM understanding, moving beyond syntactic-level metrics to assess relational and structural fidelity. S3KG's type-separated, one-to-one label alignment addresses fundamental limitations of clustering-based approaches while maintaining the WL kernel's ability to capture multi-hop neighborhood similarity.

The consistent performance gap between PubMedQA and MesaQA scores across all 3 models reveals a measurable difference in LLM capability for domain-specific versus general health knowledge. The two-dimensional S3KG evaluation further distinguishes factual accuracy (gold similarity) from contextual faithfulness (context similarity), exposing model-specific trade-offs that aggregate metrics cannot capture. The TAU additionally enables pinpointing specific reasoning failures, whether due to incorrect entity substitution, relation errors, or broader inconsistencies.


\section{Conclusion}


We presented a novel KG-based evaluation framework for assessing LLM contextual understanding in QA. By constructing canonicalized KGs from LLM outputs, gold answers, and context, and comparing them using S3KG, we move beyond syntactic-level accuracy toward verifiable, graph-theoretic comprehension measurement. S3KG achieves best-per-dataset F1 of $0.766$--$0.932$ and AUROC up to $0.973$, consistently outperforming lexical and neural baselines on KG-rich datasets while remaining competitive on short-text settings. The TAU provides interpretable, fine-grained diagnostics of reasoning failures. Evaluations on PubMedQA and MesaQA demonstrate consistent model-specific strengths and weaknesses, establishing a reproducible pipeline that can be extended to other datasets and tasks to support trustworthy AI development. In this sense, whether LLMs understand context becomes empirically testable by measuring how well their generated responses preserve the relational knowledge expressed in the reference answer and supporting context.

\section*{Limitations}
Key limitations of this work include: 
    (1)~sensitivity of similarity scores to KG extraction quality, as noisy or incomplete triplet extraction directly degrades S3KG performance; 
    (2)~computational cost of SBERT inference at scale, which may be prohibitive for very large evaluation sets without GPU acceleration; and 
    (3)~evaluation is currently restricted to open-source 7B-parameter models; extending to larger proprietary models such as GPT-4 remains future work. Additionally, the current alignment scheme does not handle directional semantic equivalence (e.g., \textit{daughter\_of} vs. \textit{mother\_of}), which would require attention-based mechanisms.

\section*{Acknowledgments}

The work of Kamal Premaratne (KP) was supported by the French/US joint project LUCAS between the Agence Nationale de la Recherche (ANR) (grant ANR-25-CE23-2189) and the U.S. National Science Foundation (NSF) (grant numbers 2530255 and 2530256).


\bibliography{references}


\appendix


\section{Effect of Temperature on Contextual Understanding}
\label{sec:appendix_temp}


Tables~\ref{tab:temp_gold} and~\ref{tab:temp_ctx} report mean GoldSim and ContextSim respectively for each model across temperature settings $T \in \{0.0, 0.3, 0.7, 1.0\}$ on both datasets. Most models show little sensitivity to temperature, with score variations within $\pm 0.01$--$0.02$ across all settings. The exception is Falcon-7B on MesaQA, where GoldSim drops substantially from $0.6036$ at $T = 0.0$ to $0.4662$ at $T = 1.0$, indicating that higher sampling randomness significantly degrades factual alignment for this model. Scores tend to peak mildly at $T = 0.3$ for most models before declining at $T = 1.0$, suggesting that a small degree of randomness can marginally improve contextual grounding without sacrificing factual accuracy. PubMedQA scores are notably more stable across temperatures than MesaQA, likely due to the constrained nature of biomedical answers.

\begin{table}[htbp]
    \centering
    \caption{Mean GoldSim per model across temperatures. CUS remains stable across temperature settings, with $T = 0.0$ serving as a reliable default for controlled evaluation.}
    \label{tab:temp_gold}
    \setlength{\tabcolsep}{4pt}
    \resizebox{\columnwidth}{!}{%
    \begin{tabular}{l l cccc}
        \toprule
        \textbf{Dataset} 
            & \textbf{Model} 
            & \textbf{$T = 0.0$} & \textbf{$T = 0.3$} & \textbf{$T = 0.7$} & \textbf{$T = 1.0$} \\
        \midrule
        {}
            & Llama-2-7B 
            & 0.6570 & \textbf{0.6588} & 0.6580 & 0.6406 \\
        MesaQA   
            & Gemma-7B   
            & 0.6889 & \textbf{0.7060} & 0.7015 & 0.6931 \\
        {}
            & Mistral-7B 
            & 0.6491 & \textbf{0.6567} & 0.6523 & 0.6444 \\
        {}
            & Falcon-7B  
            & 0.6036 & \textbf{0.6085} & 0.5558 & 0.4662 \\
        \midrule
        {}
            & Llama-2-7B 
            & \textbf{0.5220} & 0.5185 & 0.5143 & 0.5087 \\
        PubMedQA 
            & Gemma-7B   
            & \textbf{0.5235} & 0.5149 & 0.5153 & 0.5204 \\
        {}
            & Mistral-7B 
            & \textbf{0.5138} & 0.5121 & 0.5085 & 0.5023 \\
        {}
            & Falcon-7B  
            & \textbf{0.4541} & 0.4330 & 0.4184 & 0.3852 \\
        \bottomrule
    \end{tabular}}
\end{table}

\begin{table}[htbp]
    \centering
    \caption{Mean ContextSim per model across temperatures. CtxSim is more sensitive to temperature than GoldSim, yet remains stable for most models, with $T = 0.3$ yielding peak contextual faithfulness across both datasets before declining at higher temperatures.}
    \label{tab:temp_ctx}
    \setlength{\tabcolsep}{4pt}
    \resizebox{\columnwidth}{!}{%
    \begin{tabular}{l l cccc}
        \toprule
        \textbf{Dataset} 
            & \textbf{Model} 
            & \textbf{$T = 0.0$} & \textbf{$T = 0.3$} & \textbf{$T = 0.7$} & \textbf{$T = 1.0$} \\
        \midrule
        {}
            & Llama-2-7B 
            & 0.7236 & 0.7212 & \textbf{0.7248} & 0.7026 \\
        MesaQA   
            & Gemma-7B   
            & 0.7031 & \textbf{0.7217} & 0.7045 & 0.7013 \\
        {}
            & Mistral-7B 
            & 0.7356 & \textbf{0.7585} & 0.7287 & 0.7223 \\
        {}
            & Falcon-7B  
            & 0.6458 & \textbf{0.6571} & 0.6122 & 0.5195 \\
        \midrule
        {}
            & Llama-2-7B 
            & 0.6567 & 0.6563 & \textbf{0.6594} & 0.6468 \\
        PubMedQA 
            & Gemma-7B   
            & 0.6401 & \textbf{0.6540} & 0.6434 & 0.6319 \\
        {}
            & Mistral-7B 
            & 0.7331 & \textbf{0.7352} & 0.7314 & 0.7003 \\
        {}
            & Falcon-7B  
            & \textbf{0.5560} & 0.5165 & 0.5120 & 0.4571 \\
        \bottomrule
    \end{tabular}}
\end{table}


\section{KG Construction Prompt}
\label{sec:llm-prompt}


KGs are extracted using a structured chat-style prompt inspired by \citet{grapheval} and \citet{kea}. The prompt instructs the model to perform four sequential steps across all three input texts:
\begin{enumerate}
    \item \textbf{Entity detection:} Extract all named entities, concepts, attributes, quantities, dates, locations, and roles comprehensively.
    \item \textbf{Coreference resolution:} Replace all pronouns with their referent entity names, using consistent labels across all three texts.
    \item \textbf{Relation extraction:} Identify semantic relationships as simple, concise phrases, decomposing compound sentences into one triplet per fact.
    \item \textbf{Knowledge graph refinement:} Where the same entity or relation appears across multiple graphs, use the same label consistently without merging distinct facts.
\end{enumerate}

The model is instructed to return a JSON object with exactly 3 keys (\texttt{knowledge\_graph1}, \texttt{knowledge\_graph2}, and \hfill\break \texttt{knowledge\_graph3}), each containing a list of \texttt{[subject, relation, object]} triples. Few-shot examples are included in the system prompt to ground the expected output format and label consistency behaviour. The user turn specifies the three input sources explicitly:

\begin{itemize}
    \item \texttt{TEXT1} --- reference answer.
    \item \texttt{TEXT2} --- model-generated response.  
    \item \texttt{TEXT3} --- supporting context.
\end{itemize}

The complete prompt, including few-shot examples, is available at \href{https://github.com/aaivu/knowledge-xtraction}{https://github.com/aaivu/knowledge-xtraction}.
\section{TAU Diagnosing Context and Answer Deviations}
\label{sec:diagnosing}


\subsection{Dataset and Annotation}


We evaluate the TAU on a manually annotated subset of the \textbf{MessaQA} dataset, which contains general health-related questions with long form answers. For each QA instance, two KGs are constructed: (i) a gold KG extracted from the reference answer, and (ii) an LLM generated (Llama and Gamma was used)  KG extracted from the model output. The task of the triplet analysis unit is to identify semantically aligned triplet pairs between these two graphs. To obtain reliable evaluation labels, we created a gold set of aligned triplet pairs (GT $\leftrightarrow$ LLM). Alignment was independently annotated by three medical students following a fixed guideline defining semantic equivalence at the triplet level. Disagreements were resolved through adjudication.


\subsection{Annotation Quality}


Inter annotator agreement was measured over candidate aligned triplet pairs. The results indicate strong consistency, with percent agreement of 0.9433, pairwise F1 scores of 0.9708, 0.9825, and 0.9882, and pairwise Jaccard scores of 0.9433, 0.9657, and 0.9766, confirming the reliability of the annotations.


\subsection{Evaluation Protocol}


Performance is evaluated against the annotated alignments using precision, recall, and F1 score, reported using both micro averaged metrics (aggregated over all triplets) and macro averaged metrics (computed per QA instance and averaged), capturing both overall performance and consistency across samples.


\subsection{Results}


As shown in Table~\ref{tab:triplet}, the KEA baseline achieves higher precision due to its conservative component wise matching, but exhibits low recall, missing many valid alignments. In contrast, our sentence-level alignment approach significantly improves recall (+34.8\%), resulting in higher Micro and Macro F1 scores. This improvement arises from robustness to lexical variation in relations. For example, semantically equivalent relations such as \textit{``treats''} and \textit{``used for''} are often not aligned by KEA, whereas our method captures such equivalence through sentence level semantic similarity. The lower precision reflects the expected trade-off when moving from strict lexical matching to semantic matching, while the overall F1 improvement indicates a better balance between sensitivity and specificity. Additionally, our method enables residual error analysis by categorizing mismatches into relation wrong and entity wrong types.

\begin{table}[t]
\centering
\caption{Comparison of TAU alignment performance between our method and the KEA baseline, evaluated on manually annotated MesaQA and PubMed subsets.}
\label{tab:triplet}
\begin{tabular}{lcc}
\toprule
\textbf{Metric} & \textbf{KEA Baseline} & \textbf{Ours} \\
\midrule
Micro Precision & \textbf{0.853} & 0.847 \\
Micro Recall    & 0.738 & \textbf{0.949} \\
Macro Precision    & \textbf{0.953} & 0.803 \\
Macro Recall    & 0.641 & \textbf{0.946} \\
Micro F1        & 0.836 & \textbf{0.895} \\
Macro F1        & 0.622 & \textbf{0.782} \\
\bottomrule
\end{tabular}
\end{table}

\section{Per-Dataset Performance Tables}
\label{app:per_dataset}


This appendix decomposes the headline results of Table~\ref{tab:heatmap} into the 3 benchmark families, with F1 and AUROC shown in adjacent columns so that threshold-dependent and ranking behaviour can be inspected side by side. All datasets share the same protocol: $N = 400$ balanced pairs, per-method threshold sweep for F1, and AUROC on the raw scores. For S3KG, the best $\alpha$ identified in Appendix~\ref{app:alpha_sweep} is used per dataset, and the best score per column is \textbf{bolded}. The Wikipedia Entity-Swap table (Table~\ref{tab:swap}) additionally reports precision and recall to expose ROUGE-1's near-perfect score as a token-overlap artifact rather than a meaningful signal.




\begin{table}[htbp]
    \centering
    \caption{Performance on Short-Text Datasets. S3KG leads on PAWS-Wiki (F1 0.766, AUROC 0.795) but trails sentence-T5-base on MRPC and STS12, where dense semantic representations have a natural advantage over structural signals.}
    \label{tab:short}
    \setlength{\tabcolsep}{3pt}
    \resizebox{\columnwidth}{!}{%
    \begin{tabular}{l cc cc cc}
        \toprule
        {}
            & \multicolumn{2}{c}{\textbf{MRPC}}
            & \multicolumn{2}{c}{\textbf{PAWS-Wiki}}
            & \multicolumn{2}{c}{\textbf{STS12}} \\
        \cmidrule(lr){2-3} \cmidrule(lr){4-5} \cmidrule(lr){6-7}
        \textbf{Method}
            & \textbf{F1} & \textbf{AUROC} 
            & \textbf{F1} & \textbf{AUROC} 
            & \textbf{F1} & \textbf{AUROC} \\
        \midrule
        S3KG (Ours)             
            & 0.692 & 0.673
            & \textbf{0.766} & \textbf{0.795} 
            & 0.786 & 0.834 \\
        \midrule
        ROUGE-1
            & 0.745 & 0.784
            & 0.678 & 0.490
            & 0.725 & 0.754 \\
        ROUGE-2
            & 0.720 & 0.721
            & 0.715 & 0.721
            & 0.681 & 0.656 \\
        ROUGE-L
            & 0.729 & 0.760
            & 0.735 & 0.807
            & 0.703 & 0.710 \\
        BLEU
            & 0.687 & 0.677
            & 0.716 & 0.747
            & 0.671 & 0.644 \\
        BERTScore
            & 0.758 & 0.816
            & 0.691 & 0.702
            & 0.682 & 0.636 \\
        MiniLM
            & 0.723 & 0.748
            & 0.687 & 0.638
            & 0.833 & 0.894 \\
        sentence-T5-base 
            & \textbf{0.766} & \textbf{0.816} 
            & 0.674 & 0.668
            & \textbf{0.853} & \textbf{0.928} \\
        \bottomrule
    \end{tabular}}
\end{table}




ROUGE-1 achieves a near-perfect score on this dataset due to a known artifact: entity-swapped pairs differ only in the swapped entity tokens while sharing nearly identical surrounding surface-form tokens, making unigram overlap trivially high. ROUGE-1 is therefore excluded from the meaningful comparison.

\begin{table}[htbp]
    \centering
    \caption{Results on Wikipedia Entity-Swap ($N = 400$). 
    S3KG attains the highest F1 and AUROC across all retained baselines; ROUGE-1 is excluded as its near-perfect score is a token-overlap artifact rather than a meaningful signal.}
    \label{tab:swap}
    \setlength{\tabcolsep}{6pt}
    \begin{tabular}{l cc}
        \toprule
        \textbf{Method} 
            & \textbf{F1} & \textbf{AUROC} \\
        \midrule
        S3KG (Ours)  
            & \textbf{0.872} & \textbf{0.890} \\
        \midrule
        ROUGE-2
            & 0.860 & 0.772 \\
        ROUGE-L
            & 0.729 & 0.311 \\
        BLEU
            & 0.868 & 0.745 \\
        BERTScore
            & 0.747 & 0.645 \\
        MiniLM
            & 0.821 & 0.811 \\
        sentence-T5-base 
            & 0.762 & 0.806 \\
        \bottomrule
    \end{tabular}
\end{table}



\vspace*{-\baselineskip}

\begin{table*}[t]
    \centering
    \caption{Performance on KG-Perturbed Paragraph Datasets. 
    C400: SK-Codex~400; Comb.: SK-Combined; Find: SK-FindKG; Oreg.: SK-Oregano. 
    S3KG leads on four of five datasets; sentence-T5-base outperforms on SK-FindKG, the only dataset where pure dense embeddings dominate ($\alpha = 0.0$ is optimal).}
    \label{tab:kg}
    \setlength{\tabcolsep}{5pt}
    \renewcommand{\arraystretch}{1.05}
    \resizebox{\textwidth}{!}{%
    \begin{tabular}{l cc cc cc cc cc}
        \toprule
        {}
            & \multicolumn{2}{c}{\textbf{C400}}
            & \multicolumn{2}{c}{\textbf{Comb.}}
            & \multicolumn{2}{c}{\textbf{Find}}
            & \multicolumn{2}{c}{\textbf{GloBI}}
            & \multicolumn{2}{c}{\textbf{Oreg.}} \\
        \cmidrule(lr){2-3} \cmidrule(lr){4-5} \cmidrule(lr){6-7} \cmidrule(lr){8-9} \cmidrule(lr){10-11}
        \textbf{Method} 
            & \textbf{F1} & \textbf{AUROC} 
            & \textbf{F1} & \textbf{AUROC} 
            & \textbf{F1} & \textbf{AUROC} 
            & \textbf{F1} & \textbf{AUROC} 
            & \textbf{F1} & \textbf{AUROC} \\
        \midrule
        S3KG (Ours)
            & \textbf{0.932} & \textbf{0.973} 
            & \textbf{0.834} & \textbf{0.829} 
            & 0.767 & 0.796
            & \textbf{0.892} & \textbf{0.935} 
            & \textbf{0.812} & \textbf{0.892} \\
        \midrule
        ROUGE-1
            & 0.835 & 0.917
            & 0.732 & 0.728
            & 0.745 & 0.745
            & 0.784 & 0.833
            & 0.745 & 0.782 \\
        ROUGE-2
            & 0.822 & 0.894
            & 0.707 & 0.711
            & 0.717 & 0.706
            & 0.776 & 0.833
            & 0.752 & 0.791 \\
        ROUGE-L
            & 0.792 & 0.855
            & 0.722 & 0.717
            & 0.719 & 0.721
            & 0.763 & 0.800
            & 0.792 & 0.835 \\
        BLEU
            & 0.806 & 0.884
            & 0.715 & 0.708
            & 0.711 & 0.710
            & 0.775 & 0.819
            & 0.745 & 0.794 \\
        BERTScore
            & 0.823 & 0.916
            & 0.757 & 0.792
            & 0.739 & 0.761
            & 0.816 & 0.871
            & 0.743 & 0.798 \\
        MiniLM
            & 0.875 & 0.943
            & 0.770 & 0.789
            & 0.802 & 0.844
            & 0.780 & 0.817
            & 0.773 & 0.814 \\
        sentence-T5-base
        & 0.876 & 0.944
        & 0.770 & 0.827
        & \textbf{0.848} & \textbf{0.902} 
        & 0.728 & 0.760
        & 0.797 & 0.871 \\
    \bottomrule
    \end{tabular}}
\end{table*}


\section{Hyper-parameter Selection}
\label{app:alpha_sweep}


The mixing coefficient $\alpha$ in \eqref{eq:S3KG} is the sole hyper-parameter of S3KG, controlling the trade-off between the structural WL kernel signal ($\alpha = 0.0$) and the SBERT semantic signal ($\alpha = 1.0$), with intermediate values blending both. We tune $\alpha$ per dataset via grid search over $\{0.0, 0.1, \ldots, 1.0\}$, selecting the value maximising binary-classification F1; AUROC is reported alongside to confirm the result is not an artefact of threshold sensitivity.

Table~\ref{tab:alpha_sweep_full} consolidates the full sweep across all nine benchmark datasets. The best $\alpha$ per dataset (selected by maximum F1) is \textbf{bolded} together with its F1 / AUROC entries. Three patterns emerge: 
    (i)~datasets dominated by perturbations (SK-FindKG, Wiki Swap, STS12) favour low $\alpha \in \{0.0, 0.1\}$, where the WL signal carries most of the discriminative power; 
    (ii)~datasets where both relational and lexical paraphrase signals are simultaneously informative (SK-Codex~400, SK-Combined, PAWS-Wiki) peak in the balanced range $\alpha \in [0.4, 0.6]$; 
    (iii)~the AUROC surface is consistently flatter than the F1 surface, indicating that $\alpha$ primarily reshapes the score distribution near the decision boundary.

\makeatletter
\setlength{\@fptop}{0pt}
\setlength{\@dblfptop}{0pt}
\makeatother
\begin{table*}[!t]
    \centering
    \caption{Full S3KG $\alpha$ sweep across short-text and combined datasets (F1 / AUROC). The best $\alpha$ per dataset (by maximum F1) is \textbf{bolded}. Balanced datasets such as PAWS-Wiki and SK-Combined peak at $\alpha \in [0.4, 0.6]$, while STS12 favours the KG-heavy end ($\alpha = 0.1$).}
    \label{tab:alpha_sweep_full}
    \setlength{\tabcolsep}{5pt}
    \renewcommand{\arraystretch}{1.05}
    \resizebox{\textwidth}{!}{%
    \begin{tabular}{c cc cc cc cc cc}
        \toprule
        {}
            & \multicolumn{2}{c}{\textbf{MRPC}}
            & \multicolumn{2}{c}{\textbf{PAWS-Wiki}}
            & \multicolumn{2}{c}{\textbf{STS12}}
            & \multicolumn{2}{c}{\textbf{SK-Codex 400}}
            & \multicolumn{2}{c}{\textbf{SK-Combined}} \\
        \cmidrule(lr){2-3} \cmidrule(lr){4-5} \cmidrule(lr){6-7} \cmidrule(lr){8-9} \cmidrule(lr){10-11}
        $\alpha$ 
            & F1 & AUROC 
            & F1 & AUROC 
            & F1 & AUROC 
            & F1 & AUROC 
            & F1 & AUROC \\
        \midrule
        0.0 
            & 0.676 & 0.654 
            & 0.694 & 0.739 
            & 0.783 & 0.828 
            & 0.871 & 0.969 
            & 0.776 & 0.801 \\
        0.1 
            & 0.681 & 0.664 
            & 0.745 & 0.781 
            & \textbf{0.786} & \textbf{0.834} 
            & 0.922 & 0.973
            & 0.791 & 0.825 \\
        0.2     
            & 0.683 & 0.669 
            & 0.760 & 0.790 
            & 0.780 & 0.834 
            & 0.927 & 0.973 
            & 0.819 & 0.829 \\
        0.3 
            & \textbf{0.692} & \textbf{0.673} 
            & 0.764 & 0.793 
            & 0.780 & 0.831 
            & 0.927 & 0.973 
            & 0.828 & 0.830 \\
        0.4 
            & 0.680 & 0.674 
            & 0.764 & 0.795 
            & 0.780 & 0.827 
            & 0.929 & 0.973 
            & 0.833 & 0.829 \\
        0.5
            & 0.680 & 0.675 
            & \textbf{0.766} & \textbf{0.795} 
            & 0.775 & 0.824 
            & \textbf{0.932} & \textbf{0.973} 
            & \textbf{0.834} & \textbf{0.829} \\
        0.6 
            & 0.681 & 0.674 
            & 0.764 & 0.795 
            & 0.768 & 0.820 
            & 0.932 & 0.973 
            & 0.832 & 0.828 \\
        0.7 
            & 0.688 & 0.674 
            & 0.764 & 0.795 
            & 0.764 & 0.815 
            & 0.932 & 0.973 
            & 0.833 & 0.828 \\
        0.8 
            & 0.684 & 0.672 
            & 0.764 & 0.795 
            & 0.762 & 0.810 
            & 0.932 & 0.973 
            & 0.832 & 0.828 \\
        0.9 
            & 0.683 & 0.671 
            & 0.764 & 0.795 
            & 0.756 & 0.805 
            & 0.932 & 0.973 
            & 0.828 & 0.827 \\
        1.0 
            & 0.681 & 0.671 
            & 0.764 & 0.731 
            & 0.756 & 0.790 
            & 0.932 & 0.932 
            & 0.828 & 0.799 \\
        \bottomrule
    \end{tabular}}
\end{table*}

\makeatletter
\setlength{\@fptop}{0pt}
\setlength{\@dblfptop}{0pt}
\makeatother
\begin{table*}[!t]
    \caption{Full S3KG $\alpha$ sweep across KG-perturbed and entity-swap datasets (F1 / AUROC). The best $\alpha$ per dataset (by maximum F1) is \textbf{bolded}. Structure-dominated datasets (SK-FindKG, Wiki Swap) favour low $\alpha \in \{0.0, 0.1\}$, confirming the WL kernel's advantage on relational perturbations.}
    \label{tab:alpha_sweep_full_part2}
    \setlength{\tabcolsep}{9pt}
    \renewcommand{\arraystretch}{1.05}
    \resizebox{\textwidth}{!}{%
    \begin{tabular}{c cc cc cc cc}
        \toprule
        {}
            & \multicolumn{2}{c}{\textbf{SK-FindKG}}
            & \multicolumn{2}{c}{\textbf{SK-GloBI}}
            & \multicolumn{2}{c}{\textbf{SK-Oregano}}
            & \multicolumn{2}{c}{\textbf{Wiki Swap}} \\
        \cmidrule(lr){2-3} \cmidrule(lr){4-5} \cmidrule(lr){6-7} \cmidrule(lr){8-9}
        $\alpha$ 
            & F1 & AUROC 
            & F1 & AUROC 
            & F1 & AUROC 
            & F1 & AUROC \\
        \midrule
        0.0 
            & \textbf{0.767} & \textbf{0.796} 
            & 0.811 & 0.898 
            & 0.777 & 0.884 
            & 0.821 & 0.892 \\
        0.1 
            & 0.760 & 0.798 
            & 0.883 & 0.934 
            & 0.803 & 0.892 
            & \textbf{0.872} & \textbf{0.890} \\
        0.2 
            & 0.762 & 0.793 
            & 0.888 & 0.937 
            & 0.811 & 0.892 
            & 0.869 & 0.890 \\
        0.3 
            & 0.748 & 0.788 
            & 0.889 & 0.936 
            & 0.812 & 0.892 
            & 0.868 & 0.890 \\
        0.4 
            & 0.744 & 0.783 
            & 0.891 & 0.936 
            & \textbf{0.812} & \textbf{0.892} 
            & 0.865 & 0.890 \\
        0.5 
            & 0.741 & 0.779 
            & 0.890 & 0.935 
            & 0.810 & 0.892 
            & 0.868 & 0.890 \\
        0.6 
            & 0.736 & 0.776 
            & \textbf{0.892} & \textbf{0.935} 
            & 0.812 & 0.892 
            & 0.865 & 0.889 \\
        0.7 
            & 0.734 & 0.772 
            & 0.892 & 0.935 
            & 0.812 & 0.891 
            & 0.865 & 0.889 \\
        0.8 
            & 0.731 & 0.769 
            & 0.892 & 0.935 
            & 0.810 & 0.891 
            & 0.865 & 0.889 \\
        0.9 
            & 0.729 & 0.765 
            & 0.892 & 0.934 
            & 0.810 & 0.891 
            & 0.865 & 0.889 \\
        1.0 
            & 0.728 & 0.756 
            & 0.892 & 0.917 
            & 0.810 & 0.770 
            & 0.865 & 0.852 \\
        \bottomrule
    \end{tabular}}
\end{table*}



\end{document}